\documentclass[PHME, 2026]{PHMSociety}

\usepackage{graphicx}
\usepackage{amsmath,amssymb}
\usepackage{booktabs}
\usepackage{multirow}
\usepackage{subcaption}
\makeatletter\providecommand\@xviiipt{17.28}\providecommand\@xxiiipt{23}\makeatother 
\usepackage{enumitem}
\usepackage{tikz}
\usetikzlibrary{positioning, arrows.meta, calc}
\usepackage{xcolor}

\usepackage{makecell}
\usepackage{array}

\newcommand{\rev}[1]{#1} 

\title{%
  Spatiotemporal Graph Neural Networks for Fault Detection and \\
  Structural Learning in Chemical Processes: Use Case on the Tennessee Eastman Process%
}

\author{%
  Rayane Ammar Khodja\authorNumber{1,2},
  Alexandre Voisin\authorNumber{2},
  Victor Costa\authorNumber{1},
  Fanny Casteran\authorNumber{1},
  Benoit Celse\authorNumber{1}
  \And Beno\^{i}t Iung\authorNumber{2}
}

\address{%
    \affiliation{{1}}{IFP Energies Nouvelles, Rond-point de l'\'{e}changeur de Solaize, BP 3, 69360 Solaize, France}{
        {\email{rayane.ammar-khodja@ifpen.fr}}\\
		{\email{victor.costa@ifpen.fr}}\\
		{\email{fanny.casteran@ifpen.fr}}\\
		{\email{benoit.celse@ifpen.fr}}
	}
    \tabularnewline
	\affiliation{{2}}{Universit\'{e} de Lorraine, CNRS, CRAN, F-54000 Nancy, France}{
		{\email{alexandre.voisin@univ-lorraine.fr}}\\
		{\email{benoit.iung@univ-lorraine.fr}}
	}
	
}

\begin{document}

\maketitle
\pagestyle{fancy}
\thispagestyle{plain}

\maketitle
\pagestyle{fancy}
\thispagestyle{plain}

\phmLicenseFootnote{Rayane Ammar Khodja}

\begin{abstract}%
Classical Fault Detection and Diagnosis (FDD) methods, including many
data-driven approaches, assume a static normal operating space and
interpret deviations from a fixed reference as fault indicators.  When
Operating Conditions (OCs) vary over time, this assumption breaks
down: legitimate transitions trigger false alarms while moderate
faults go undetected.  We propose a spatiotemporal Graph Neural
Network (GNN) framework that decomposes the normal operating space
into OC-specific subspaces linked by transition functions, with a
dual learning objective combining reconstruction loss and a Deep
Support Vector Data Description (DeepSVDD) one-class term.  The
framework learns adjacency matrices through Graph Attention Networks
(GATs) and integrates spatial modelling with temporal encoding to
represent process dynamics under evolving OCs.

This paper evaluates the foundational components of the
framework, fault detection, spatial graph learning via GATv2 and temporal encoding, on the Tennessee Eastman Process (TEP)
benchmark, with training performed exclusively on fault-free data. The spatiotemporal architecture
achieves competitive detection performance from reconstruction error
alone, with similarity-based feature selection improving both accuracy
and graph structure diversity.  We then evaluate the physical
interpretability of the learned attention matrices against 22
ground-truth sensor pairs derived from the TEP control structure and
process topology. 
The GATv2 attention does not recover all the necessary known physical
pairs across multiple hyperparameter configurations, suggesting a
structural limitation of reconstruction-driven attention rather than
a tuning issue.
This result challenges a common assumption in GNN-based FDD:
that learned attention weights provide a basis for fault diagnosis and
root-cause analysis.  The architecture detects faults effectively, but
the learned graph does not encode the physical topology needed for
interpretable diagnosis, motivating physics-informed graph
construction.
\end{abstract}


\section{Introduction}
\label{sec:intro}

Fault Detection and Diagnosis (FDD) in chemical processes is critical
for ensuring safety, product quality, and economic
efficiency~\cite{venkatasubramanian2003a}.  Modern industrial systems
generate high-dimensional multivariate time-series data from tens to
hundreds of sensors.  Data-driven methods, including Principal
Component Analysis (PCA)~\cite{chiang2001}, autoencoders
\cite{sakurada2014}, and Long Short-Term Memory (LSTM)
networks~\cite{filonov2016}, have been widely adopted for anomaly
detection on such data.

Nevertheless, most existing approaches treat the sensor array as a flat
feature vector, ignoring the physical topology of the process (including PCA-based monitoring
\cite{chiang2001}, deep autoencoders~\cite{sakurada2014},
LSTM-based predictive models~\cite{filonov2016}, and the broader
data-driven FDD literature reviewed by~\citeA{alauddin2018}).  In
reality, sensors are embedded in a network of material flows, energy
balances, and control loops that govern how disturbances propagate
through the plant.  Graph Neural Networks (GNNs) offer a natural
framework for encoding these structural relationships, with nodes
representing sensors and edges capturing inter-sensor
dependencies~\cite{wu2021}. By propagating information along graph edges that reflect physical process connections, GNNs can exploit the known topology of the plant
to improve both detection sensitivity and diagnostic
interpretability.

A critical gap in this literature is the absence of systematic
evaluation of whether graph structures learned by attention mechanisms
reflect physically meaningful sensor relationships, or merely optimize
for the downstream reconstruction objective.  Most studies feed raw
sensor data to GAT-based models and assume, without verification, that
the resulting attention weights capture real process physics.  This question is especially relevant for process industries, where
interpretability of the learned model is essential for operator
trust and root-cause analysis.  We review existing GNN-based FDD
approaches and their treatment of this assumption in
Section~\ref{sec:related_work}.


In this paper, we address this gap directly.  We build on the
MTAD-GAT architecture of~\citeA{zhao2020}, which is a dual feature--temporal
graph attention model originally proposed for infrastructure
anomaly detection (details in Section~\ref{sec:related_work}), and adapt it to the
chemical-process setting, where, to the best of our knowledge, it
has not previously been evaluated.  Around this backbone, we
present a spatiotemporal GNN framework for FDD under dynamic
operating conditions and evaluate its foundational components
(spatial graph learning and temporal encoding) and fault detection on the TEP
benchmark~\cite{downs1993,rieth2017} under fixed operating
conditions.  The OC-aware extensions, including dual-loss training
and multi-mode evaluation, are the subject of ongoing work and
outlined in Section~\ref{sec:conclusion}.  Our contributions are:
\begin{enumerate}[leftmargin=*,itemsep=2pt]
    \item The first application of the MTAD-GAT architecture
        \cite{zhao2020} to a chemical process benchmark (TEP).  This
        transfer required five methodological modifications to the
        original training and evaluation pipeline, enumerated in
        Section~\ref{sec:method};
  \item A systematic evaluation of the physical interpretability of
    GATv2~\cite{brody2021} attention weights using a Hit-Rate@$k$ metric against
    process-knowledge-derived ground truth;
  \item An empirical analysis of the gap between reconstruction-driven
    attention and physical process topology, motivating physics-informed
    graph construction.
\end{enumerate}

This paper is organised as follows. Section~\ref{sec:related_work} surveys GNN-based FDD methods and their treatment of graph construction and physical interpretability.
Section~\ref{sec:framework} presents the proposed framework, including
the OC-aware formulation, the dual-loss objective, and the
spatiotemporal architecture used to learn the sensor graph.
Section~\ref{sec:method} details the experimental protocol, covering
the Tennessee Eastman Process benchmark, feature selection, the data
pipeline, anomaly scoring, ground-truth physical pairs, and
evaluation metrics.  Section~\ref{sec:results} reports detection
performance, graph structure evaluation and the limitations of our work.  Section~\ref{sec:conclusion}
draws final conclusions and outlines directions for future work.


\section{Related Work}
\label{sec:related_work}
\rev{
This section reviews graph neural network methods for fault detection
and diagnosis in industrial processes, organized around the two
choices that most directly bear on the present work: how the inter-
sensor graph is constructed, and whether the detection objective is
reconstruction- or prediction-based.  Comprehensive overviews of the
broader GNN-FDD landscape are given by~\citeA{chen2024} and the recent
review of GNN applications in process modelling. 

The earliest GNN-based methods for chemical process FDD relied on
graphs derived from process knowledge. \citeA{wu2021ptcn} proposed the
Process Topology Convolutional Network (PTCN), which feeds a Graph
Convolutional Network \cite{kipf2017} with an adjacency matrix
constructed from the plant's piping-and-instrumentation diagram
(P\&ID).  \citeA{jia2023} formalized this idea as Topology-Guided
Graph Learning (TGGL): a Signed Directed Graph (SDG) prior is
injected into the GCN, and the model is biased toward learned
structures consistent with known causal relationships.  These
methods report strong fault classification performance on the
Tennessee Eastman Process (TEP), but require either a verified
process topology or expert-curated causal graphs.

An additional set of works treat the adjacency matrix itself as a
trainable parameter.  \citeA{deng2021} introduced the Graph Deviation
Network (GDN) for multivariate time-series anomaly detection,
learning sensor relations from a low-rank embedding rather than from
prior knowledge.  \citeA{kovalenko2024} extended this idea by
allowing multiple parallel adjacency matrices to be optimized
end-to-end on TEP, reporting that learned graphs can match or
exceed knowledge-derived ones for fault classification.\\
\citeA{chen2023} proposed Interaction-Aware Graph Neural Networks
(IAGNN), in which a heterogeneous graph with multiple edge types is
constructed, and edge weights are learned by attention.  These
trainable-graph methods relax the dependence on expert knowledge,
but leave open the question raised by~\citeA{jia2023}: whether the
resulting graphs encode physically meaningful structure or merely
statistical co-variation.

Graph Attention Networks~\cite{velickovic2018,brody2021} have
attracted particular interest because the attention coefficients
$\alpha_{ij}$ admit an interpretation as edge weights, making the
learned graph nominally inspectable.  \citeA{zhao2020} proposed
MTAD-GAT, a dual feature--temporal graph attention architecture for
multivariate anomaly detection, originally validated on
infrastructure benchmarks (SMD, MSL, SMAP) rather than chemical
processes.  \citeA{liu2024} combined a CNN feature extractor with a
GAT layer guided by Granger-causality maps for TEP fault detection
and root-cause analysis.  Across these works, the attention matrix
is treated as a structural artifact that supports diagnostic
interpretation, yet none provides a systematic evaluation of
whether the attention weights recover the underlying process
topology.

GNN-based detectors fall into two methodological families.
Prediction-based approaches, such as GDN~\cite{deng2021} and
CNN-GAT~\cite{liu2024}, train the model to forecast the next
timestep and flag deviations between predictions and observations.
Reconstruction-based approaches, including
MTAD-GAT~\cite{zhao2020}, train an autoencoder to reproduce the
input window and flag deviations between input and reconstruction.
Both paradigms can in principle drive the learning of a sensor
graph through their respective gradients, but they impose different
inductive biases: reconstruction emphasises which sensors are
predictive of which others within the current window, while
prediction emphasises which sensors lead which others in time.
Our framework follows the reconstruction paradigm because it aligns
with the OC-aware decomposition (Section~\ref{sec:framework}), in
which each operating condition defines a static normal manifold
rather than a dynamical regime; the structural question we ask is
whether reconstruction-driven attention recovers physical topology
on TEP.

Most of the works above do not evaluate the physical interpretability of
the learned attention or adjacency matrix against an explicit
process-knowledge ground truth.  Several explicitly assume such
interpretability without verifying
it~\cite{zhao2020,liu2024,chen2023}; others demonstrate that
physics-informed graphs outperform purely data-driven ones for
classification~\cite{jia2023,wu2021ptcn} but do not isolate the
question of whether attention alone can recover the prior.  The
present work targets this gap.
}
\section{Proposed Framework}
\label{sec:framework}

Most reconstruction-based fault detection methods aggregate all training data
into a single model of normal behaviour \\
~\cite{alauddin2018}. The resulting normal space must
accommodate every operating condition the plant has seen, inflating the
decision boundary to encompass the full operational envelope.  Abrupt
faults of moderate magnitude fall inside this inflated boundary and go
undetected; progressive faults such as catalyst deactivation or sensor
drift are detected only after the deviation has accumulated well beyond
the point where early intervention was possible.  The method functions
as an extreme-event detector rather than a sensitive fault monitor.
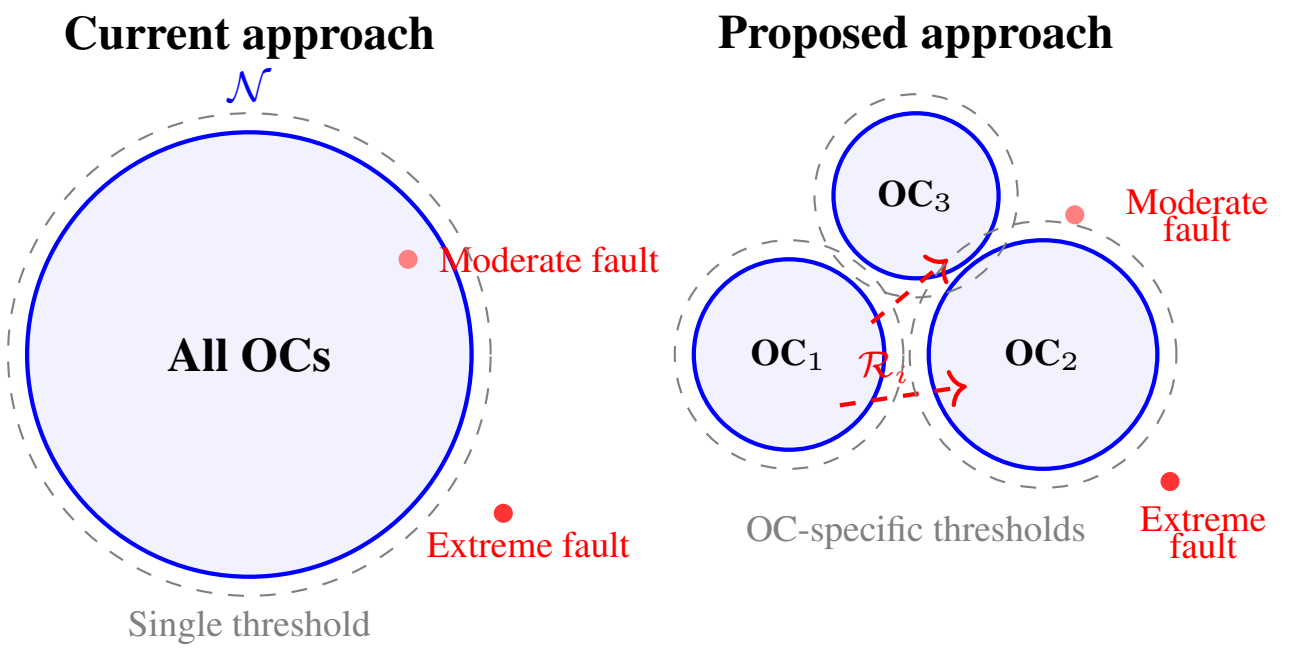
\begin{figure}[htbp]
  \centering
  \begin{tikzpicture}[scale=0.42]
    \begin{scope}[xshift=-6.5cm]
      \draw[thick, blue, fill=blue!5] (0,0) circle (3.5);
      \node[font=\small\bfseries] at (0,0) {All OCs};
      \node[blue, font=\footnotesize, above] at (0,3.7) {$\mathcal{N}$};
      \fill[red!50] (2.5,1.5) circle (0.15);
      \node[red, font=\tiny, right] at (2.7,1.5) {Moderate fault};
      \fill[red!80] (4,-2.5) circle (0.15);
      \node[red, font=\tiny, right] at (2.5,-3) {Extreme fault};
      \draw[dashed, gray] (0,0) circle (3.8);
      \node[gray, font=\tiny] at (0,-4.3) {Single threshold};
      \node[font=\small\bfseries] at (0,5) {Current approach};
    \end{scope}
    \begin{scope}[xshift=4cm]
      \draw[thick, blue, fill=blue!5] (-2,0) circle (1.5);
      \draw[thick, blue, fill=blue!5] (2,0) circle (1.8);
      \draw[thick, blue, fill=blue!5] (0,2.5) circle (1.3);
      \node[font=\tiny\bfseries] at (-2,0) {OC$_1$};
      \node[font=\tiny\bfseries] at (2,0) {OC$_2$};
      \node[font=\tiny\bfseries] at (0,2.5) {OC$_3$};
      \draw[->, thick, red, dashed] (-0.7,0.5) -- (0.5,1.5);
      \draw[->, thick, red, dashed] (-1.2,-0.8) -- (0.8,-0.5);
      \node[red, font=\tiny] at (-0.5,-0.2) {$\mathcal{R}_i$};
      \fill[red!50] (2.5,2.2) circle (0.15);
      \node[red, font=\tiny, right] at (2.7,2.2)
        {\parbox{1.2cm}{\centering Moderate\\[-2pt]fault}};
      \fill[red!80] (4,-2) circle (0.15);
      \node[red, font=\tiny, right] at (2.8,-2.8)
        {\parbox{1.2cm}{\centering Extreme\\[-2pt]fault}};
      \draw[dashed, gray] (-2,0) circle (1.8);
      \draw[dashed, gray] (2,0) circle (2.1);
      \draw[dashed, gray] (0,2.5) circle (1.6);
      \node[gray, font=\tiny] at (0,-2.8) {OC-specific thresholds};
      \node[font=\small\bfseries] at (0,5) {Proposed approach};
    \end{scope}
  \end{tikzpicture}
  \caption{Single-threshold detection (left) versus OC-aware detection
    (right).}
  \label{fig:oc_comparison}
\end{figure}

Figure~\ref{fig:oc_comparison} illustrates the problem.  Under a
single-threshold approach (left), intentional OC transitions widen the
normal space $\mathcal{N}$ until only faults of extreme magnitude
exceed the boundary.  An OC-aware decomposition (right) partitions
$\mathcal{N}$ into condition-specific subspaces
$\{\mathcal{N}_{OC_1}, \ldots, \mathcal{N}_{OC_K}\}$, each with its
own tighter boundary, so that moderate faults become distinguishable
from normal variability within any given condition.


\begin{figure*}[h!]
  \centering
  \includegraphics[width=0.75\linewidth]{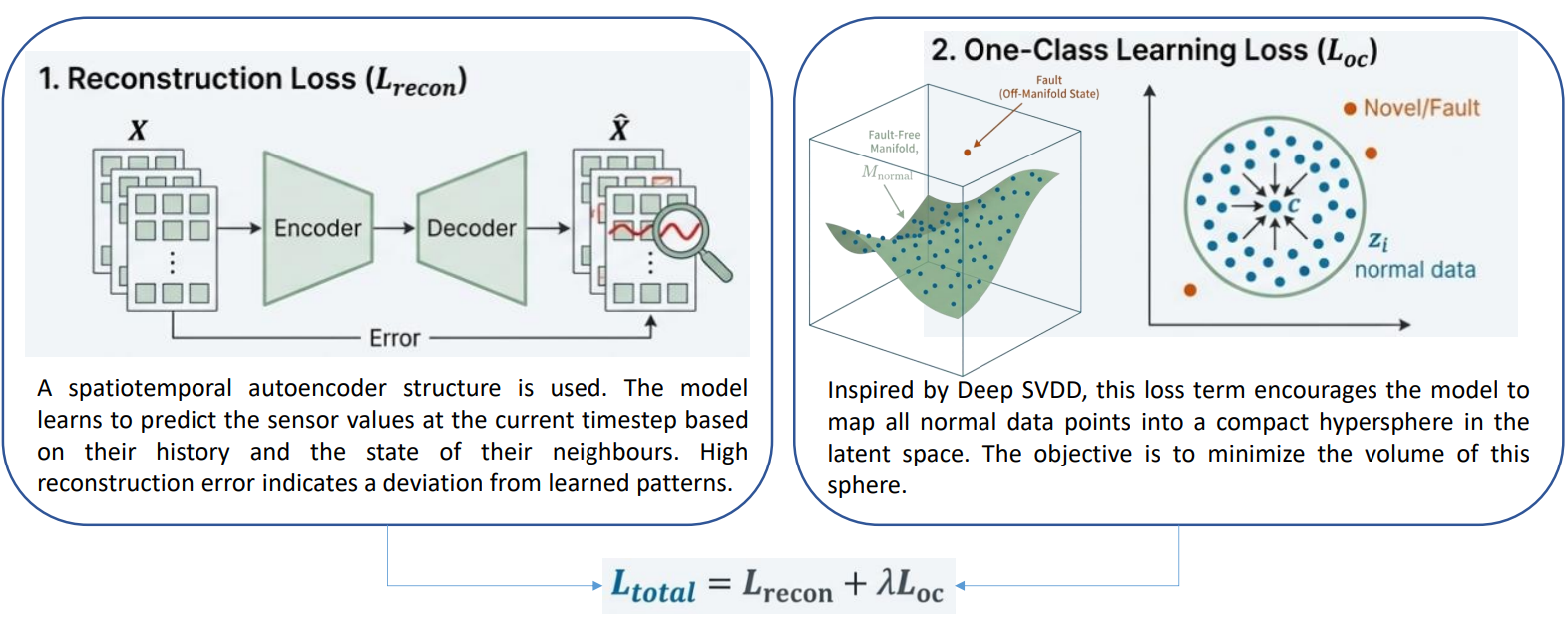}
  \caption{Dual-loss framework.  Left: the spatiotemporal Graph Autoencoder (GAE) for reconstruction.  Right: a Deep SVDD-inspired one-class loss maps
    normal data to a compact hypersphere in latent space.}
  \label{fig:dual_loss}
\end{figure*}

Formally, we define normal behaviour not as a single space but as a
collection of OC-specific subspaces linked by transformation functions:
\begin{equation}
  \mathcal{N} = \bigl\{\mathcal{N}_{\text{prep}},\;
    \mathcal{N}_{OC_1},\; \ldots,\;
    \mathcal{N}_{OC_K}\bigr\}
  \label{eq:oc_decomposition}
\end{equation}
\begin{equation}
  \mathcal{R}_i : \mathcal{N}_{OC_i} \rightarrow
    \mathcal{N}_{OC_{i+1}}
  \label{eq:transition}
\end{equation}
This decomposition enables four distinguishable system states:
(1)~\textit{preparation}: readings inside $\mathcal{N}_{\text{prep}}$
as the plant is brought to its initial operating point;
(2)~\textit{normal within an OC}: readings inside
$\mathcal{N}_{OC_i}$; (3)~\textit{normal transition}: evolution
following $\mathcal{R}_i$ between adjacent subspaces; and
(4)~\textit{actual fault}: deviations outside all OC boundaries,
preparation envelope, and transition corridors.

The framework combines two complementary objectives (Figure~\ref{fig:dual_loss}).  A spatiotemporal autoencoder
reconstructs sensor values from their history and neighbourhood
context; high reconstruction error signals deviation from learned
normal patterns.  A Deep SVDD-inspired one-class loss term~\cite{ruff2018}
simultaneously maps normal data into a compact hypersphere in latent
space, tightening the decision boundary beyond what reconstruction
error alone provides.  

The model consumes the multivariate
sensor stream as a sequence of overlapping sliding windows.  Let
$\mathbf{X} \in \mathbb{R}^{L \times N}$ denote a recording of length
$L$ over $N$ sensors, where $\mathbf{x}_t \in \mathbb{R}^{N}$ is the
sensor vector at timestep $t$.  A window of length $T$ starting at
index $w$ is
\begin{equation}
  X_w = \bigl[\mathbf{x}_{w},\, \mathbf{x}_{w+1},\, \ldots,\,
              \mathbf{x}_{w+T-1}\bigr]^{\top}
        \in \mathbb{R}^{T \times N},
  \label{eq:window}
\end{equation}
with successive windows separated by a stride $s$.  During training, a
batch of $B$ such windows is processed jointly, producing the input
tensor of shape $(B, T, N)$.  The model reconstructs the last
timestep of each window, $\hat{\mathbf{x}}_{w,T} \in \mathbb{R}^{N}$,
from the spatial and temporal context provided by the preceding
$T-1$ steps.  All loss terms below operate on this windowed
representation.

For a single window $X_w \in \mathbb{R}^{T \times N}$
(Eq.~\ref{eq:window}), the reconstruction loss measures the squared
deviation between the actual and reconstructed sensor values at the
last timestep $t = T$:

\begin{equation}
  L_{\text{recon}} = \frac{1}{N}\sum_{i=1}^{N}
    \bigl(x_{T,i} - \hat{x}_{T,i}\bigr)^{2}
  \label{eq:recon_loss}
\end{equation}

where $x_{T,i}$ and $\hat{x}_{T,i}$ denote the observed and
reconstructed values of sensor $i$ at the last timestep of the window
$X_w$. The total training loss averages
$L_{\text{recon}}$ over all windows in a batch.

The one-class loss is calculated as follows:
\begin{equation}
  L_{\text{oc}} = \frac{1}{M}\sum_{j=1}^{M}
    \bigl\|\phi(\mathbf{x}^{(j)};\, \mathcal{W}) - \mathbf{c}\bigr\|^{2}
  \label{eq:oc_loss}
\end{equation}
where $\mathbf{x}^{(j)} \in \mathbb{R}^{T \times N}$ is the $j$-th
fault-free training window (Eq.~\ref{eq:window}),
$\phi(\cdot;\, \mathcal{W})$ is the encoder network with parameters
$\mathcal{W}$, $M$ is the number of fault-free training windows, and
$\mathbf{c}$ is the hypersphere centre computed as the mean of initial
encoder outputs on fault-free data~\cite{ruff2018}.

The total training objective combines both terms:
\begin{equation}
  L_{\text{total}} = L_{\text{recon}} + \lambda\, L_{\text{oc}}
  \label{eq:dual_loss}
\end{equation}

where $\lambda$ controls the relative weight of the one-class term.
The current paper evaluates the reconstruction component \\
($L_{\text{recon}}$); integrating the one-class term is part of
ongoing work.

The remainder of this paper focuses exclusively on the reconstruction
branch of the framework, i.e spatiotemporal graph autoencoder on the
left of Figure~\ref{fig:dual_loss}, trained under
$L_{\text{recon}}$ alone, and evaluates it under fixed operating
conditions using the modified MTAD-GAT backbone described above.  The
central empirical question we examine is whether the GATv2 attention
matrix learned through reconstruction alone recovers physically
meaningful sensor relationships, or whether reconstruction-driven
optimisation alone is insufficient to surface the process topology.
The OC-aware decomposition
(Eqs. ~\ref{eq:oc_decomposition} and \ref{eq:transition}), the one-class
loss term $L_{\text{oc}}$, and dynamic-OC evaluation are deferred
to future work and revisited in Section~\ref{sec:conclusion}.


\subsection{Spatial Graph Learning}


At each timestep, the $N$ sensor channels of a window
$X_w \in \mathbb{R}^{T \times N}$ (Eq.~\ref{eq:window}) are treated as
the nodes of a fully connected directed graph
$\mathcal{G} = (\mathcal{V}, \mathcal{E})$ with
$|\mathcal{V}| = N$.  Each node $i$ is associated with a feature
vector $\mathbf{v}_i \in \mathbb{R}^{T}$ that contains its $T$ recent
observations within the window.  The graph is fully connected to
avoid imposing prior structure on the learning problem: the
attention mechanism is asked to discover which pairs of sensors are
mutually informative from data alone.

We use GATv2 rather than the original GAT~\cite{velickovic2018} because the latter
computes only \emph{static} attention: the ranking of source nodes
attended to by a given target is fixed by the learned parameters and
does not depend on the target's own features.  \citeA{brody2021} show
that this restriction prevents the original GAT from representing
certain simple node-ranking problems.

For each ordered pair $(i, j)$ of sensors, GATv2~\cite{brody2021}
computes an attention coefficient $\alpha_{ij}$ that quantifies how
much sensor $j$'s features contribute to the updated representation
of sensor $i$.  The coefficient is obtained in two steps.  First, a
shared learnable linear projection $W$ and a learnable attention
vector $\mathbf{a}$ produce a scalar score for each pair:
\begin{equation}
  e_{ij} = \mathbf{a}^{\top}\,
           \mathrm{LeakyReLU}\!\bigl(
              W\,[\mathbf{v}_i \,\Vert\, \mathbf{v}_j]
           \bigr),
  \label{eq:gatv2_score}
\end{equation}
where $\Vert$ denotes vector concatenation.  Second, the scores are
normalised across all neighbours of $i$ via a softmax:
\begin{equation}
  \alpha_{ij} = \frac{\exp(e_{ij})}
                     {\sum_{k=1}^{N} \exp(e_{ik})}.
  \label{eq:gatv2_softmax}
\end{equation}
The updated node representation is a weighted sum of neighbour
features:
$\mathbf{v}_i' = \sigma\bigl(\sum_{j} \alpha_{ij}\, W\,\mathbf{v}_j\bigr)$,
where $\sigma$ is a non-linearity.  Stacking the $\alpha_{ij}$
yields a dense adjacency matrix
$A \in \mathbb{R}^{N \times N}$ whose rows sum to one.

\subsection{Temporal Encoding}
\begin{figure}[t]
  \centering
  \resizebox{0.75\linewidth}{!}{%
    \begin{tikzpicture}[
        node distance=0.3cm,
        block/.style={rectangle, draw, rounded corners=2pt,
          minimum height=0.55cm, minimum width=1.8cm,
          font=\scriptsize\bfseries, align=center, thick},
        input/.style={block, fill=blue!8, draw=blue!50},
        conv/.style={block, fill=orange!12, draw=orange!50},
        attn/.style={block, fill=green!10, draw=green!50!black},
        rnn/.style={block, fill=purple!10, draw=purple!50},
        outblk/.style={block, fill=red!8, draw=red!40},
        arr/.style={->, thick, >=stealth},
        lbl/.style={font=\tiny, text=black!60, align=left, xshift=0.2cm}
      ]
        \node[input] (inp) {Input $(B, T, N)$};
        \node[conv, below=of inp] (conv) {1-D ConvLayer};
        \node[attn, below=of conv] (feat) {Feature Attention (GATv2)};
        \node[attn, below=of feat] (temp) {Temporal Attention};
        \node[rnn, below=of temp] (gru) {GRU Encoder};
        \node[outblk, below=of gru] (dec) {Linear Decoder};
        \node[input, below=of dec] (outp) {$\hat{x}_{t=T}$};

        \draw[arr] (inp) -- (conv);
        \draw[arr] (conv) -- (feat);
        \draw[arr] (feat) -- (temp);
        \draw[arr] (temp) -- (gru);
        \draw[arr] (gru) -- (dec);
        \draw[arr] (dec) -- (outp);

        \draw[arr, rounded corners=4pt]
          (conv.east) -- ++(+1, 0) |- (temp.east);

        \node[lbl, anchor=east, xshift=-0.15cm] at (feat.west) {inter-sensor};
        \node[lbl, anchor=east, xshift=-0.15cm] at (temp.west) {timestep};
        \node[lbl, anchor=east, xshift=-0.15cm] at (gru.west) {sequence};

        \draw[arr, dashed, red!60, rounded corners=6pt]
          (outp.east) -- ++(0.9, 0) |- node[pos=0.1, rotate=-90,
            yshift=0.15cm, font=\tiny, text=red!60]
            {anomaly score} (inp.east);

    \end{tikzpicture}%
  }
  \caption{Architecture of the spatiotemporal graph attention
    autoencoder.  The 1-D ConvLayer provides local temporal smoothing.
    Feature attention (GATv2) learns inter-sensor relationships;
    temporal attention models timestep dependencies.  Both outputs are
    concatenated and encoded by a GRU.  A linear decoder reconstructs
    $\hat{x}_{t=T}$; reconstruction error serves as the anomaly
    score.}
  \label{fig:architecture}
\end{figure}
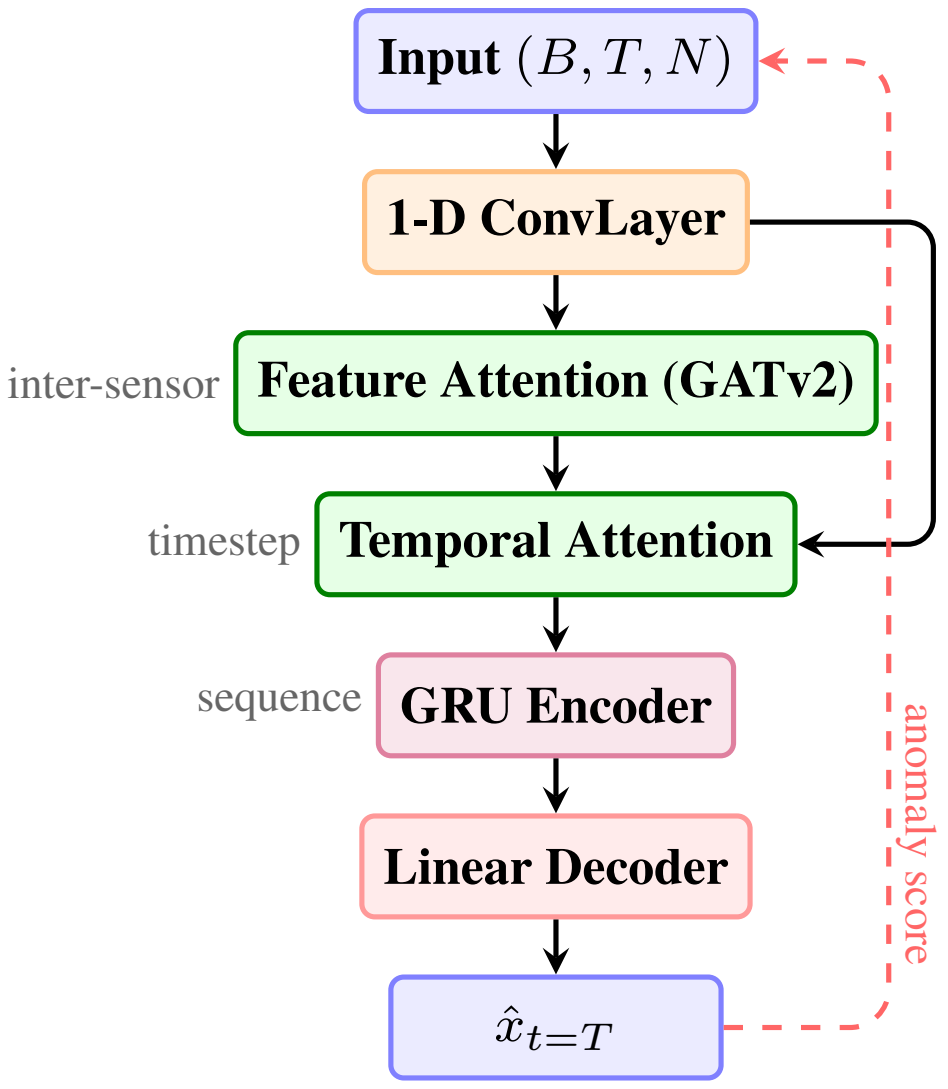

A Gated Recurrent Unit (GRU) layer encodes the
temporal dynamics of the concatenated spatial and temporal attention
outputs.  We chose the GRU over the LSTM for two reasons.  First, the GRU's reduced parameter count (no separate cell-state and output gates) is generally associated with faster convergence and reduced overfitting risk when training data is limited, motivating its use in this setting.
Second, the spatial and temporal attention layers upstream already
provide multiplicative gating over the input sequence, which reduces
the marginal value of the additional gating mechanisms an LSTM
provides.  A linear decoder reconstructs the last timestep of each
input window from the GRU's final hidden state.  The anomaly score
is the per-feature mean squared error between reconstructed and
actual values.  Figure~\ref{fig:architecture} summarises the complete
architecture.

\section{Experiment}
\label{sec:method}

\subsection{Case Study: Tennessee Eastman Process}
\label{sec:tep}

The Tennessee Eastman Process (TEP), introduced by
\citeA{downs1993}, is a realistic simulation of a continuous chemical
plant comprising five major unit operations: a two-phase reactor, a
condenser, a vapor--liquid separator, a recycle compressor, and a
reboiled stripper.  The process involves eight chemical components, four
reactants (A, C, D, E), two products (G, H), one byproduct (F), and one
inert (B), undergoing exothermic, irreversible reactions.

The system tracks 52 variables: 22 continuous process measurements
($xmeas_1$--$xmeas_{22}$) covering pressures, temperatures, levels, and flow
rates; 19 composition analyser variables ($xmeas_{23}$--$xmeas_{41}$) sampled at
6--15 minute intervals with corresponding dead times; and
11 manipulated variables ($xmv_1$--$xmv_{11}$) representing valve positions
and flow setpoints.

We use the extended dataset of \citeA{rieth2017}, which provides
500 simulation runs of 500 samples each (25 hours at 3-minute
intervals) for fault-free training, and 500 runs of 960 samples each
(48 hours) for each of the 20 fault types.  The fault scenarios span
four disturbance mechanisms: step changes in feed conditions and
temperatures (Faults~1--7), random variations in feed and cooling water
parameters \\
(Faults~8--12), a slow drift in reaction kinetics
(Fault~13), valve sticking (Faults~14--15), and five undisclosed
disturbances (Faults~16--20).  Faults are introduced at sample~161
(8 hours into test simulations) after an initial fault-free period.
Complete variable definitions and fault descriptions are provided in
the Appendix A.

\subsection{Feature Selection}
\label{sec:feature_selection}

Following \citeA{jia2023}, we apply similarity-based feature selection
to remove variables with low discriminative information across fault
types.  For each variable $v$ and fault type $z$, a similarity
coefficient is computed:
\begin{equation}
  r_{z,v} = 1 - \frac{\|\mathbf{x}_{z,v} - \mathbf{x}_{0,v}\|_2}
    {\sum_{z=1}^{Z}\|\mathbf{x}_{z,v} - \mathbf{x}_{0,v}\|_2 +
    \varepsilon}
  \label{eq:similarity}
\end{equation}
where $\mathbf{x}_{0,v}$ is the normal-condition signal and
$\varepsilon$ prevents division by zero.  Variables whose coefficient
exceeds a threshold across all fault types are removed, reducing the
feature set from 52 to 35 variables. The variables removed include 
composition analyzers with slow sampling rates
($xmeas_{28}$--$xmeas_{41}$), compressor work ($xmeas_{20}$) and two
manipulated variables ($xmv_5$, $xmv_9$).

\subsection{Data Pipeline}

The preprocessing pipeline enforces strict separation between
training, validation, and test data to prevent information leakage
and ensure that evaluation reflects genuine out-of-sample performance.

\textbf{Normalisation.} A \texttt{StandardScaler} is fitted
exclusively on fault-free training data.  The same fitted scaler is
applied, without re-fitting, to validation and test splits, so that
no distributional information from faulty or held-out data influences
the scaling parameters.

\textbf{Data partitioning.} The train/validation split operates at
the simulation-run level: each of the 500 fault-free runs (500 samples
per run) is assigned entirely to either training (400 runs) or
validation (100 runs).  This prevents adjacent sliding windows drawn
from the same run from appearing in both sets, which would create
temporal leakage.  Faulty data is reserved exclusively for testing.

\textbf{Attention weight extraction.} The GATv2 feature attention
layer stores attention coefficients directly after the softmax
operation, ensuring that the extracted weights correspond to the
actual pairwise attention distribution used during message passing.

\textbf{Convolutional preprocessing.} A 1-D convolutional layer
provides local temporal smoothing before the input reaches both the
feature and temporal attention layers.  Its output is fed directly
into the attention modules.

\subsection{Within-Window Detrending}

To reduce the sensitivity of GATv2 attention to absolute signal levels,
we apply within-window mean subtraction:
\begin{equation}
  \tilde{X}_{w} = X_{w} - \frac{1}{T}\sum_{t=1}^{T} X_{w,t}
  \label{eq:detrend}
\end{equation}
where $X_w \in \mathbb{R}^{T \times N}$ is a single window.  This
forces the attention mechanism to operate on relative signal variations
rather than absolute operating-point values.

\subsection{Anomaly Scoring}

The model is trained as a reconstruction autoencoder on fault-free
data.  At test time, the anomaly score for each window is the
per-feature mean squared error between reconstructed and actual last
timestep:
\begin{equation}
  s_w = \frac{1}{N}\sum_{i=1}^{N}(x_{w,T,i} - \hat{x}_{w,T,i})^{2}
  \label{eq:anomaly_score}
\end{equation}
Windows from fault-free test data yield low scores; faulty windows
produce elevated scores. 

\subsection{Ground-Truth Physical Pairs}
\label{sec:ground_truth}

To evaluate whether the learned attention matrix encodes process
structure, we compile 22 ground-truth sensor pairs from three sources
of TEP process knowledge:

\begin{enumerate}[leftmargin=*,itemsep=2pt]
  \item \textbf{Control loops} (11 pairs): Direct sensor--actuator
    feedback connections from the decentralised control
    structure~\cite{downs1993};
  \item \textbf{Material flows} (8 pairs): Sensors connected by
    physical streams (feed, reactor--separator, separator--stripper);
  \item \textbf{Thermodynamic couplings} (3 pairs): Energy-balance
    and ideal-gas-law relationships within process units.
\end{enumerate}

These pairs are validated against the SDG analysis of
\citeA{maurya2004} and the bond graph model of
\citeA{tidriri2018}.  Evaluation uses the Hit-Rate@$k$ metric: the
fraction of the top-$k$ learned or correlation-based pairs that match
the ground truth.  Pearson correlation (absolute value) serves as a
competing baseline for graph construction (full list in
Appendix B).

\subsection{Evaluation Metrics}
\label{sec:metrics}

We evaluate the framework along two axes: fault detection
discrimination and graph structure interpretability.

\textbf{ROC-AUC.} The Receiver Operating Characteristic curve plots
the true positive rate against the false positive rate as the
decision threshold varies.  The area under this curve quantifies
how well the anomaly score separates faulty from fault-free windows:
\begin{equation}
  \text{ROC-AUC} = \int_0^1 \text{TPR}\bigl(\text{FPR}^{-1}(t)\bigr)\, dt
  \label{eq:rocauc}
\end{equation}
A score of 1.0 indicates perfect separation; 0.5 corresponds to
random ranking.

\textbf{PR-AUC.} The Precision--Recall curve plots precision against
recall.  Its area is more informative than ROC-AUC when the class
distribution is imbalanced, as is the case when fault-free windows
substantially outnumber faulty ones:
\begin{equation}
  \text{PR-AUC} = \int_0^1 \text{Precision}\bigl(\text{Recall}^{-1}(t)\bigr)\, dt
  \label{eq:prauc}
\end{equation}
A high PR-AUC confirms that elevated anomaly scores correspond
predominantly to true faults rather than false alarms.

\textbf{Hit-Rate@$k$.} To assess whether the learned adjacency
matrix encodes physically meaningful structure, we rank all
$N(N-1)$ directed sensor pairs by their attention weight (for GAT)
or absolute Pearson correlation (for the baseline) and compute
the fraction of the top-$k$ pairs that match the 22 ground-truth
physical pairs:
\begin{equation}
  \text{Hit-Rate@}k = \frac{|\mathcal{P}_k \cap \mathcal{G}|}{k}
  \label{eq:hitrate}
\end{equation}
where $\mathcal{P}_k$ is the set of top-$k$ ranked pairs and
$\mathcal{G}$ is the ground-truth set.  A high hit-rate indicates
that the ranking assigns strong weights to pairs with known physical
coupling; a low hit-rate suggests that the ranking reflects
reconstruction utility rather than process topology.

\subsection{Experimental Setup}
\label{sec:setup}

Table~\ref{tab:hyperparams} summarizes the hyperparameter
configurations evaluated.  All experiments share: GRU hidden
dimension 128, 2 GRU layers, temporal attention embedding
dimension 64, stride 25, batch size 64, learning rate $10^{-3}$ with scheduling of 
\texttt{ReduceLROnPlateau} , gradient clipping at norm 1.0,
and early stopping with patience 15.

\begin{table}[h!]
\small
\centering
\caption{Hyperparameter configurations of experiments (\texttt{kernel}=7).}
\label{tab:hyperparams}
\begin{tabular}{@{}lcccc@{}}
\toprule
Exp. & \texttt{feat\_embed} & \texttt{window}  & \texttt{dropout} \\
\midrule
A & 16 & 50  & 0.2 \\
B & 32 & 50  & 0.2 \\
C & 64 & 25  & 0.2 \\
D & 32 & 25  & 0.3 \\
E & 64 & 50  & 0.4 \\
\bottomrule
\end{tabular}
\end{table}


Each experiment trains a model on fault-free data (400 simulation runs
for training, 100 for validation) and evaluates on 20 per-fault test
sets, each combining fault-free test windows (label 0) with one fault
type's windows (label 1).  We report ROC-AUC and PR-AUC per fault and
averaged across all 20 faults, along with Hit-Rate@$k$ at
$k = 10, 20, 30, 50$ for both GAT attention and Pearson correlation
against the 22 ground-truth pairs.


\section{Results}
\label{sec:results}

\subsection{Fault Detection Performance}
\label{sec:results_detection}

Table~\ref{tab:detection} presents the mean detection performance
across all 20 TEP faults.  All window-size-50 configurations achieve
comparable performance (mean ROC-AUC $\approx 0.77$, mean PR-AUC
$\approx 0.80$), indicating that detection capability is robust to the
feature attention embedding dimension and dropout rate.  Smaller windows
($T = 25$) reduce performance slightly, consistent with the loss of
temporal context for the GRU encoder. Per-fault detection performance for Experiment~A is detailed in Table ~\ref{tab:per_fault}; we show the results of this experiment since it gives the best results.

\begin{table}[h!]
\small
\centering
\caption{Mean detection performance across 20 TEP faults (\texttt{f\_embed} is feature embedding, \texttt{w} is the window size and \texttt{drop} is the dropout rate).}
\label{tab:detection}
\begin{tabular}{@{}lccccc@{}}
\toprule
Exp. & \texttt{f\_embed} & \texttt{w} & \texttt{drop} & ROC-AUC & PR-AUC \\
\midrule
A & 16 & 50 & 0.2 & 0.772 & 0.803 \\
B & 32 & 50 & 0.2 & 0.771 & 0.803 \\
C & 64 & 25 & 0.2 & 0.752 & 0.784 \\
D & 32 & 25 & 0.3 & 0.754 & 0.786 \\
E & 64 & 50 & 0.4 & 0.771 & 0.802 \\
\bottomrule
\end{tabular}
\end{table}

\begin{table}[h!]
\small
\centering
\caption{Per-fault detection performance for Experiment~A. Faults marked $^{\dagger}$ are widely regarded as undetectable in the TEP literature and are routinely excluded from comparison studies.}
\label{tab:per_fault}
\begin{tabular}{@{}clcc@{}}
\toprule
F. & Type & ROC & PR \\
\midrule
1              & Step   & 0.700 & 0.769 \\
2              & Step   & 0.640 & 0.696 \\
3$^{\dagger}$  & Step   & 0.500 & 0.500 \\
4              & Step   & 0.515 & 0.545 \\
5              & Step   & 0.633 & 0.702 \\
6              & Step   & 0.912 & 0.938 \\
7              & Step   & 0.695 & 0.767 \\
8              & Random & 0.927 & 0.954 \\
9$^{\dagger}$  & Random & 0.537 & 0.527 \\
10             & Random & 0.900 & 0.931 \\
11             & Random & 0.887 & 0.921 \\
12             & Random & 0.931 & 0.956 \\
13             & Drift    & 0.918 & 0.948 \\
14             & Sticking & 0.918 & 0.947 \\
15$^{\dagger}$ & Sticking & 0.519 & 0.516 \\
16             & Unknown  & 0.804 & 0.833 \\
17             & Unknown  & 0.920 & 0.949 \\
18             & Unknown  & 0.893 & 0.927 \\
19             & Unknown  & 0.802 & 0.811 \\
20             & Unknown  & 0.885 & 0.919 \\
\midrule
\multicolumn{4}{@{}l@{}}{\footnotesize Mean (all 20 faults): ROC-AUC $0.772$, PR-AUC $0.803$} \\
\multicolumn{4}{@{}l@{}}{\footnotesize Mean (excl. F3, F9, F15): ROC-AUC $0.816$, PR-AUC $0.854$} \\
\bottomrule
\end{tabular}
\end{table}

Table~\ref{tab:lit_comparison} situates our detection performance
against published TEP results for representative reconstruction- and
graph-based methods.  Direct numerical comparison is constrained by
heterogeneous evaluation protocols across the literature: studies
differ in the train/test split (fault-free run counts, sample-level
versus run-level partitioning), the subset of faults reported (most
work excludes Faults~3, 9, and 15, which are widely regarded as
undetectable, while we include all 20), the anomaly-score
aggregation (per-sample versus windowed), and the metric definition
(some report fault detection rate at a fixed false alarm rate rather
than ROC-AUC).  Our 20-fault mean ROC-AUC of 0.77 and PR-AUC of
0.80 fall within the range reported for comparable
reconstruction-driven graph methods on TEP.

\begin{table}[h!]
\small
\centering
\caption{Reported TEP detection performance for representative
  methods. Protocols differ across studies; values are
  reproduced as published.}
\label{tab:lit_comparison}
\begin{tabular}{@{}lcc@{}}
\toprule
Method & Faults & Reported metric \\
\midrule
TGGL~\cite{jia2023}         & 20/20 & FDR $\approx 0.83$ \\
CNN-GAT~\cite{liu2024}      & 20/20 & FDR $\approx 0.85$ \\
\midrule
\textbf{This work (Exp. A)} & 20/20 & \makecell{ROC-AUC 0.77 \\ PR-AUC 0.80} \\
\bottomrule
\multicolumn{3}{@{}l@{}}{\footnotesize  TEP F1 is taken from a re-implementation reported by~\citeA{jia2023}.} \\
\multicolumn{3}{@{}l@{}}{\footnotesize FDR = Fault Detection Rate at fixed false alarm rate.} \\
\end{tabular}
\end{table}




\subsection{Graph Structure Evaluation}
\label{sec:results_adj}

Table~\ref{tab:hitrate} reports the Hit-Rate@$k$ for GAT attention and
Pearson correlation against the 22 physical ground-truth pairs, for
Experiment~A (other configurations yield similar results).

\begin{table}[h!]
\small
\centering
\caption{Hit-Rate@$k$: fraction of top-$k$ pairs matching physical
ground truth (22 pairs total).  GAT = learned attention;
Pearson = $|\rho_{ij}|$ baseline.}
\label{tab:hitrate}
\begin{tabular}{@{}rcccc@{}}
\toprule
$k$ & GAT hits & GAT rate & Pearson hits & Pearson rate \\
\midrule
10 & 0 & 0.0\% & 4 & 40.0\% \\
20 & 1 & 5.0\% & 4 & 20.0\% \\
30 & 2 & 6.7\% & 6 & 20.0\% \\
50 & 2 & 4.0\% & 7 & 14.0\% \\
\bottomrule
\end{tabular}
\end{table}

Pearson correlation consistently recovers more physical pairs than GAT
attention at all values of $k$.  The six pairs recovered by Pearson in
the top-30 are direct control-loop connections with near-unit linear
correlation (e.g., $xmeas_{12} \leftrightarrow xmv_7$,
$xmeas_{15} \leftrightarrow xmv_8$,
$xmeas_{17} \leftrightarrow xmv_{11}$), the strongest and most obvious
structural relationships in the TEP.

The GAT recovers only two physical pairs
($xmeas_7 \leftrightarrow xmv_6$ and $xmeas_3 \leftrightarrow xmv_2$), both
representing feed--valve connections. Table~\ref{tab:sweep_hitrate} examines whether this is sensitive to hyperparameter choice.
Whether the same effect persists across alternative GNN variants
(e.g., the original GAT~\cite{velickovic2018}, GCN~\cite{kipf2017})
remains an open question: the failure mode we observe is consistent
with the gradient pathology of reconstruction-driven attention
identified by~\citeA{brody2021} for the static-attention case, but
the GATv2 layer used here is strictly more expressive than the
original GAT, and the two may differ in their failure modes on this
benchmark.

Table~\ref{tab:sweep_hitrate} confirms that this result is consistent
across all hyperparameter configurations: the maximum hit count at
$k = 30$ is 1 regardless of embedding dimension, window size, or
dropout rate, indicating that the result is consistent across hyperparameter configurations rather than driven by any particular setting.

\begin{table}[h!]
\small
\centering
\caption{Hit-Rate@30 across hyperparameter configurations.}
\label{tab:sweep_hitrate}
\begin{tabular}{@{}lccc@{}}
\toprule
Exp. & Config. & Hits@30 & Rate \\
\midrule
A & 16 / 50 / 0.2 & 1 & 3.3\% \\
B & 32 / 50 / 0.2 & 1 & 3.3\% \\
C & 64 / 25 / 0.2 & 0 & 0.0\% \\
D & 32 / 25 / 0.3 & 0 & 0.0\% \\
E & 64 / 50 / 0.4 & 1 & 3.3\% \\
\bottomrule
\end{tabular}
\end{table}

\subsection{Limitations}

This study has several limitations.  First, the TEP is a simulated
benchmark; real processes present additional challenges including sensor
noise, missing data, and more complex operating condition transitions.
Second, evaluation is conducted under fixed operating conditions; testing
under dynamic conditions is deferred to future work.  Third, the
ground-truth pairs are derived from process knowledge rather than a
formally verified causal model.  Fourth, we evaluate a single base
architecture; other GNN variants may exhibit different
attention-topology relationships.

\section{Conclusion and Perspectives}
\label{sec:conclusion}
This paper presented a spatiotemporal GNN framework for fault
detection under dynamic operating conditions and evaluated its spatial graph learning via GATv2 attention, temporal encoding on the Tennessee
Eastman Process benchmark.  Our findings indicate that:
\begin{enumerate}[leftmargin=*,itemsep=2pt]
  \item The adapted MTAD-GAT architecture achieves competitive fault
    detection (mean ROC-AUC 0.77, mean PR-AUC 0.80) when trained
    exclusively on fault-free data, with performance robust to
    hyperparameter variations;
  \item The learned GATv2 attention does \emph{not} encode physically
    meaningful process topology;
  \item This effect is robust across the hyperparameter space of the
    GATv2 architecture and is consistent with reconstruction-driven
    attention emphasising predictive utility over physical
    coupling; whether the same effect persists across alternative
    GNN variants is the subject of ongoing work.
\end{enumerate}
These results establish that graph learning with spatiotemporal autoencoders is robust for fault detection, but that
the learned graph structure is not a reliable proxy for physical process
topology.  This motivates several directions for future work.  First,
implementing the hybrid adjacency matrix
$A = \alpha A_p + (1-\alpha)S$ with an SDG-derived physical prior to
anchor graph structure in known process physics.  Second, incorporating
a Deep SVDD dual-loss objective~\cite{ruff2018} to tighten the boundary
of the normal operating space beyond what reconstruction loss alone
provides.  Third, evaluating performance under dynamic operating
conditions using real industrial dataset.  Fourth, developing two-level causal
interpretability combining GNN Explainer for edge-level subgraph
extraction, to advance from detection toward root-cause diagnosis.

\section*{Nomenclature}

\begin{tabular}{ l  l }
  $N$           & Number of sensors (features) \\
  $T$           & Temporal window size (timesteps) \\
  $B$           & Batch size \\
  $\alpha_{ij}$ & GATv2 attention coefficient (sensor $j$ to $i$) \\
  $A$           & Learned adjacency matrix, $A \in \mathbb{R}^{N \times N}$ \\
  $A_p$         & Physics-derived adjacency matrix \\
  $S$           & Data-driven adjacency component \\
  $s_w$         & Window-level anomaly score \\
  $r_{z,v}$     & Similarity coefficient (fault $z$, variable $v$) \\
  $\hat{x}$     & Reconstructed observation \\
\end{tabular}

\section*{Acknowledgement}

This work is supported by IFP Energies Nouvelles and the Research
Centre for Automatic Control of Nancy (CRAN, UMR CNRS 7039),
Universit\'{e} de Lorraine.

\bibliographystyle{apacite}
\PHMbibliography{phme2026}

\clearpage
\section*{Appendix A: Tennessee Eastman Process Variables and Fault Descriptions}
\label{sec:appendix_tep}

\begin{table}[h!]
    \centering
    \small
    \caption{TEP continuous process measurements (Vertical Layout).}
    \label{tab:tep_measurements_vertical}
    \begin{tabular}{@{}llc@{}}
        \toprule
        \textbf{Var.} & \textbf{Name} & \textbf{Unit} \\
        \midrule
        \texttt{xmeas\_1}  & A Feed (stream 1)              & kscmh      \\
        \texttt{xmeas\_2}  & D Feed (stream 2)              & kg/hr      \\
        \texttt{xmeas\_3}  & E Feed (stream 3)              & kscmh      \\
        \texttt{xmeas\_4}  & A and C Feed (stream 4)        & kscmh      \\
        \texttt{xmeas\_5}  & Recycle Flow (stream 8)        & kscmh      \\
        \texttt{xmeas\_6}  & Reactor Feed Rate (stream 6)   & kscmh      \\
        \texttt{xmeas\_7}  & Reactor Pressure               & kPa gauge  \\
        \texttt{xmeas\_8}  & Reactor Level                  & \%         \\
        \texttt{xmeas\_9}  & Reactor Temperature            & °C         \\
        \texttt{xmeas\_10} & Purge Rate (stream 9)          & kscmh      \\
        \texttt{xmeas\_11} & Product Sep Temp               & °C         \\
        \texttt{xmeas\_12} & Product Sep Level              & \%         \\
        \texttt{xmeas\_13} & Prod Sep Pressure              & kPa gauge  \\
        \texttt{xmeas\_14} & Prod Sep Underflow (stream 10) & m\textsuperscript{3}/hr \\
        \texttt{xmeas\_15} & Stripper Level                 & \%         \\
        \texttt{xmeas\_16} & Stripper Pressure              & kPa gauge  \\
        \texttt{xmeas\_17} & Stripper Underflow (stream 11) & m\textsuperscript{3}/hr \\
        \texttt{xmeas\_18} & Stripper Temperature           & °C         \\
        \texttt{xmeas\_19} & Stripper Steam Flow            & kg/hr      \\
        \texttt{xmeas\_20} & Compressor Work                & kW         \\
        \texttt{xmeas\_21} & Reactor CW Outlet Temp         & °C         \\
        \texttt{xmeas\_22} & Separator CW Outlet Temp       & °C         \\
        \bottomrule
        \multicolumn{3}{@{}l@{}}{\footnotesize \textit{Note:} kscmh = thousand standard cubic metres per hour.} \\
    \end{tabular}
\end{table}

\begin{table}[h!]
    \centering
    \small
    \caption{TEP composition analyser variables (Unit is in \%mol).}
    \label{tab:tep_analysers}
    \begin{tabular}{@{}
        >{\raggedright\arraybackslash}p{2cm}
        >{\raggedright\arraybackslash}p{3cm}
        >{\raggedright\arraybackslash}p{1.5cm}
        >{\raggedright\arraybackslash}p{1.5cm}
        @{}}
        \toprule
        \textbf{Variables} & \textbf{Description} & \textbf{Sampling (h)} & \textbf{Dead Time (h)} \\
        \midrule
        \texttt{xmeas\_23} to \texttt{xmeas\_29} & Reactor Feed Analysis (Stream 6) & 0.1  & 0.1  \\
        \midrule
        \texttt{xmeas\_30} to \texttt{xmeas\_37} & Purge Gas Analysis (Stream 9)    & 0.1  & 0.1  \\
        \midrule
        \texttt{xmeas\_38} to \texttt{xmeas\_41} & Product Analysis (Stream 11)     & 0.25 & 0.25 \\
        \bottomrule
    \end{tabular}
\end{table}

\begin{table}[h!]
    \centering
    \small
    \caption{TEP manipulated variables.}
    \label{tab:tep_manipulated}
    \begin{tabular}{@{}llc@{\hspace{0.6em}}llc@{}}
        \toprule
        \textbf{Var.} & \textbf{Name} & \textbf{Unit} &
        \textbf{Var.} & \textbf{Name} & \textbf{Unit} \\
        \midrule
        \texttt{xmv\_1} & D Feed Flow (stream 2)        & kg/hr & \texttt{xmv\_7}  & Sep Pot Liquid Flow (stream 10)    & m\textsuperscript{3}/hr \\
        \texttt{xmv\_2} & E Feed Flow (stream 3)        & kg/hr & \texttt{xmv\_8}  & Stripper Product Flow (stream 11)  & m\textsuperscript{3}/hr \\
        \texttt{xmv\_3} & A Feed Flow (stream 1)        & kscmh & \texttt{xmv\_9}  & Stripper Steam Valve               & \%                      \\
        \texttt{xmv\_4} & A and C Feed Flow (stream 4)  & kscmh & \texttt{xmv\_10} & Reactor Cooling Water Flow         & m\textsuperscript{3}/hr \\
        \texttt{xmv\_5} & Compressor Recycle Valve      & \%    & \texttt{xmv\_11} & Condenser Cooling Water Flow       & m\textsuperscript{3}/hr \\
        \texttt{xmv\_6} & Purge Valve (stream 9)        & \%    &                  &                                    &                         \\
        \bottomrule
    \end{tabular}
\end{table}

\clearpage
\begin{table}[h!]
    \centering
    \small
    \caption{TEP fault descriptions and types.}
    \label{tab:tep_faults}
    \begin{tabular}{@{}clc@{\hspace{1em}}clc@{}}
        \toprule
        \textbf{F.} & \textbf{Description} & \textbf{Type} &
        \textbf{F.} & \textbf{Description} & \textbf{Type} \\
        \midrule
        1  & A/C feed ratio, B comp. (Str. 4)   & Step    & 10 & C feed temp. (Str. 4)           & Random     \\
        2  & B comp., A/C ratio (Str. 4)         & Step    & 11 & Reactor CW inlet temp.          & Random     \\
        3  & D feed temp. (Str. 2)               & Step    & 12 & Condenser CW inlet temp.        & Random     \\
        4  & Reactor CW inlet temp.              & Step    & 13 & Reaction kinetics               & Slow Drift \\
        5  & Condenser CW inlet temp.            & Step    & 14 & Reactor CW valve                & Sticking   \\
        6  & A feed loss (Str. 1)                & Step    & 15 & Condenser CW valve              & Sticking   \\
        7  & C header pressure loss (Str. 4)     & Step    & 16 & Unknown                         & ---        \\
        8  & A, B, C comp. (Str. 4)              & Random  & 17 & Unknown                         & ---        \\
        9  & D feed temp. (Str. 2)               & Random  & 18--20 & Unknown                      & ---        \\
        \bottomrule
    \end{tabular}
\end{table}

\section*{Appendix B: Ground-Truth Physical Pairs}
\label{sec:appendix_pairs}

Table~\ref{tab:ground_truth_pairs} lists the 22 ground-truth sensor
pairs used in the Hit-Rate@$k$ evaluation
(Section~\ref{sec:ground_truth}).  

\begin{table}[h!]
\small
\centering
\caption{The 22 ground-truth physical sensor pairs used for
  Hit-Rate@$k$ evaluation, grouped by source of process knowledge.}
\label{tab:ground_truth_pairs}
\begin{tabular}{@{}cllc@{}}
\toprule
\# & Sensor A & Sensor B & Description \\
\midrule
\multicolumn{4}{@{}l@{}}{\textit{Category 1 --- Control loops (11 pairs)}} \\
\midrule
1  & \texttt{xmeas\_7}  & \texttt{xmv\_4}    & Reactor pressure / A+C feed flow \\
2  & \texttt{xmeas\_7}  & \texttt{xmv\_6}    & Reactor pressure / purge valve \\
3  & \texttt{xmeas\_8}  & \texttt{xmv\_10}   & Reactor level / reactor CW flow \\
4  & \texttt{xmeas\_9}  & \texttt{xmv\_7}    & Reactor temperature / separator liquid flow \\
5  & \texttt{xmeas\_10} & \texttt{xmv\_1}    & Purge rate / D feed flow \\
6  & \texttt{xmeas\_11} & \texttt{xmv\_2}    & Product sep.\ temperature / E feed flow \\
7  & \texttt{xmeas\_12} & \texttt{xmv\_7}    & Product sep.\ level / separator liquid flow \\
8  & \texttt{xmeas\_15} & \texttt{xmv\_8}    & Stripper level / stripper product flow \\
9  & \texttt{xmeas\_17} & \texttt{xmv\_11}   & Stripper underflow / condenser CW flow \\
10 & \texttt{xmeas\_13} & \texttt{xmv\_3}    & Product sep.\ pressure / A feed flow \\
11 & \texttt{xmeas\_18} & \texttt{xmeas\_8}  & Stripper temperature / reactor level \\
\midrule
\multicolumn{4}{@{}l@{}}{\textit{Category 2 --- Material flow connections (8 pairs)}} \\
\midrule
12 & \texttt{xmeas\_1}  & \texttt{xmv\_3}    & A feed measurement / A feed flow setpoint \\
13 & \texttt{xmeas\_2}  & \texttt{xmv\_1}    & D feed measurement / D feed flow setpoint \\
14 & \texttt{xmeas\_3}  & \texttt{xmv\_2}    & E feed measurement / E feed flow setpoint \\
15 & \texttt{xmeas\_4}  & \texttt{xmeas\_6}  & A+C feed / reactor feed rate \\
16 & \texttt{xmeas\_6}  & \texttt{xmeas\_7}  & Reactor feed rate / reactor pressure \\
17 & \texttt{xmeas\_9}  & \texttt{xmeas\_11} & Reactor temp.\ / product sep.\ temperature \\
18 & \texttt{xmeas\_11} & \texttt{xmeas\_15} & Product sep.\ temp.\ / stripper level \\
19 & \texttt{xmeas\_10} & \texttt{xmeas\_18} & Purge rate / stripper temperature \\
\midrule
\multicolumn{4}{@{}l@{}}{\textit{Category 3 --- Thermodynamic couplings (3 pairs)}} \\
\midrule
20 & \texttt{xmeas\_9}  & \texttt{xmeas\_8}  & Reactor temperature / reactor level (energy balance) \\
21 & \texttt{xmeas\_7}  & \texttt{xmeas\_8}  & Reactor pressure / reactor level (ideal-gas law) \\
22 & \texttt{xmeas\_21} & \texttt{xmeas\_8}  & Reactor CW outlet temp.\ / reactor level (heat transfer) \\
\bottomrule
\end{tabular}
\end{table}


\clearpage
\section*{Appendix C: Feature-Attention Matrices}
\label{sec:appendix_gat}

\begin{figure}[htbp]
  \centering
  \includegraphics[width=0.95\textwidth]{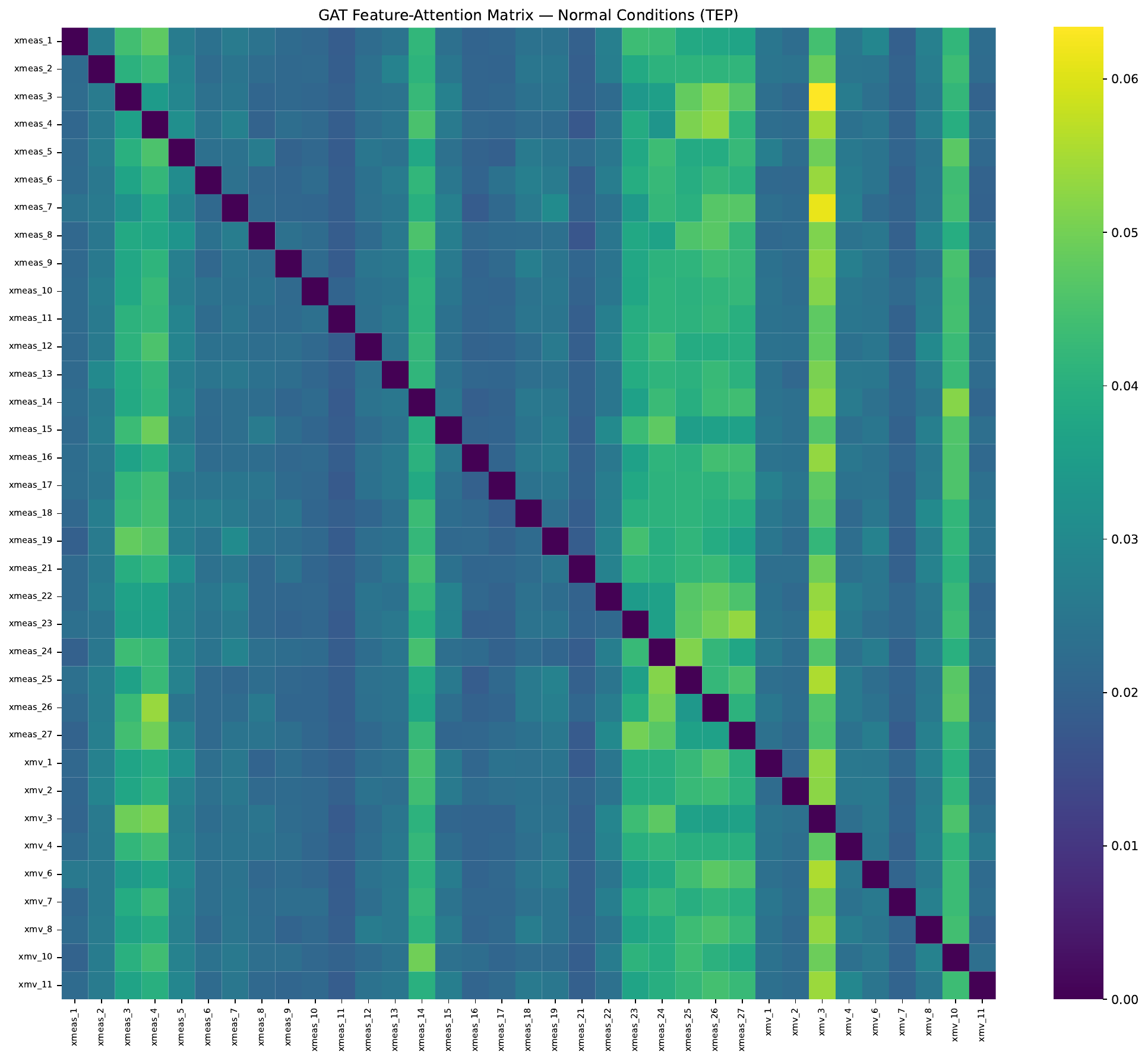}
  \caption{GATv2 feature-attention matrix learned under normal
    operating conditions (35 features, Experiment~A).  Bright vertical stripes indicate hub columns — sensors that receive high attention from many source nodes.}
  \label{fig:adjacency_normal}
\end{figure}

\begin{figure*}[htbp]
  \centering
  \includegraphics[width=1.05\textwidth]{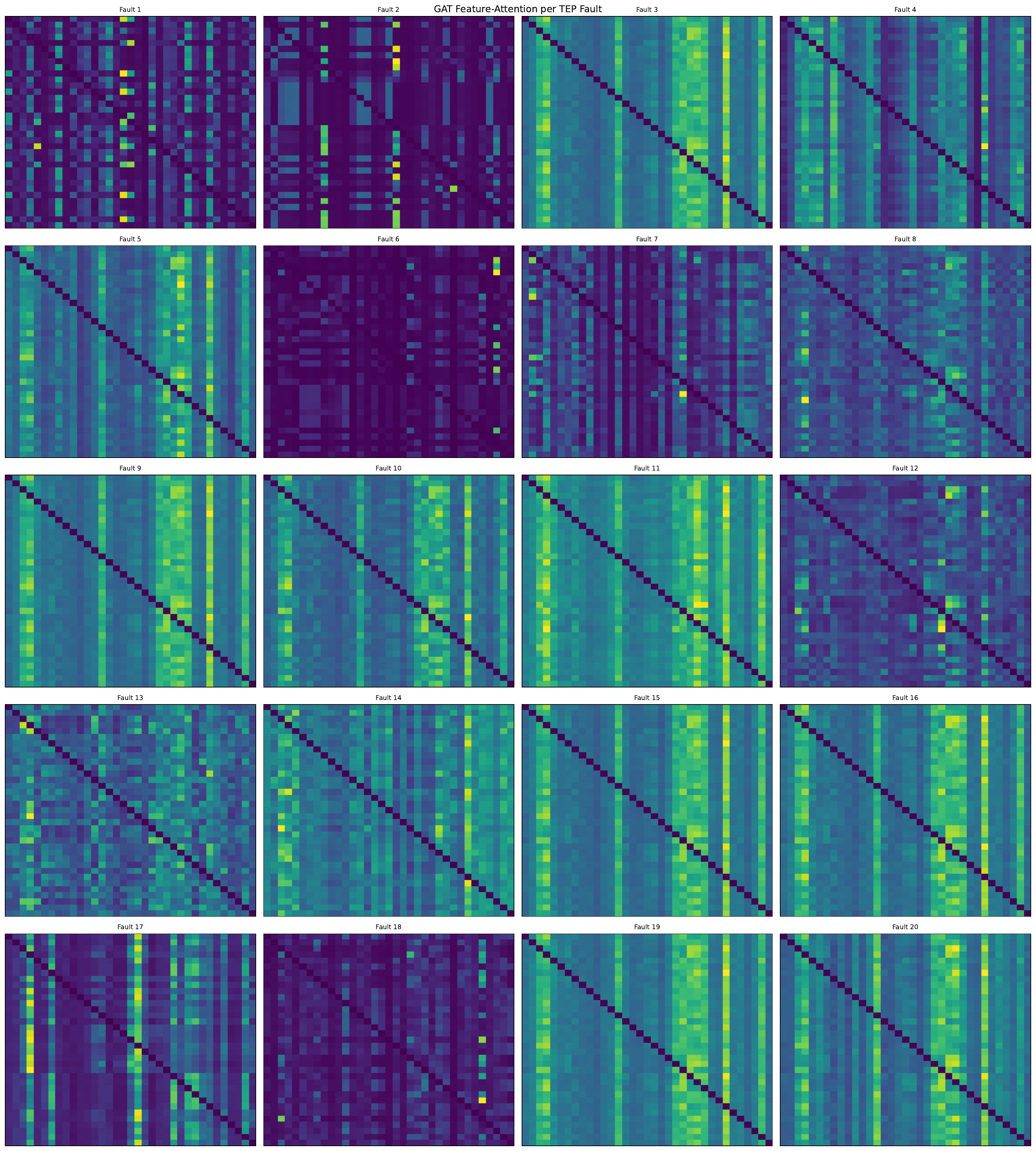}
  \caption{GATv2 feature-attention matrices computed on each of the
    20 TEP fault scenarios (Faults~1--20, left-to-right, top-to-bottom).
    Each panel shares the same axes, ordering, and colour scale as
    Figure~\ref{fig:adjacency_normal}: rows and columns index the
    35 selected sensors.}
  \label{fig:adjacency_faults}
\end{figure*}
\end{document}